\documentclass{article}
\usepackage{spconf,amsmath,graphicx,color}

\title{Towards Visually Grounded Sub-Word Speech Unit Discovery}
\name{David Harwath and James Glass}
\address{MIT Computer Science and Artificial Intelligence Laboratory\\
Cambridge, Massachusetts, 02139, USA\\
\texttt{dharwath@csail.mit.edu, glass@mit.edu}}

\begin{document}
%
\maketitle
\begin{abstract}
In this paper, we investigate the manner in which interpretable sub-word speech units emerge within a convolutional neural network model trained to associate raw speech waveforms with semantically related natural image scenes. We show how diphone boundaries can be superficially extracted from the activation patterns of intermediate layers of the model, suggesting that the model may be leveraging these events for the purpose of word recognition. We present a series of experiments investigating the information encoded by these events.  
\end{abstract}
%

%


\section{Introduction and Prior Work}
The cornerstone of automatic speech recognition (ASR) systems is the taxonomy of discrete, linguistic units modeled by the recognizer. The most salient of these is the vocabulary of words that can be recognized, but ``under the hood'' many ASR systems utilize a compositional hierarchy of units, e.g. words are composed of phonemes, and phonemes are composed of senone states. This hierarchy of linguistic units is advantageous because it offers flexibility (new words may be specified in the lexicon in terms of existing phonetic models) and data efficiency (phonetic models can be re-used across many different words, allowing for a large degree of parameter sharing between word models). However, it comes at a cost: the training data must be transcribed in terms of the acoustic units, and the compositional mapping between units (e.g. the lexicon mapping phonemes to words) must be specified in advance by an expert linguist. These annotations are expensive to collect, especially for less widely spoken, ``low-resource'' languages. Unsupervised or weakly-supervised approaches to ASR often attempt to address this problem via the automatic, data-driven discovery of these linguistic units. Some proposed approaches operate at the word level \cite{park_glass_sdtw, jansen_2010, kamper_2017}, while a separate line of work is concerned with sub-word modeling \cite{lee_glass_2012, varadarajan_2008, ondel_2016, jansen_2013a}. Other works have jointly learned sub-word as well as word-level units in a unified framework \cite{gish_2009, lee_odonnell_glass_2015}.

A central difficulty faced by unsupervised models of speech is the fact that the acoustic speech waveform is the result of a complex entanglement of many different sources of variability, such as speaker, background noise, reverberation, microphone characteristics, etc. \textit{Self-supervised} models \cite{deSa_19914} have recently garnered increased attention as an alternative approach to the traditional supervised-unsupervised dichotomy. In lieu of labels, self-supervised learning algorithms leverage informative context found e.g., in another modality. An early example of this is the CELL model introduced by \cite{roy_2002} which learned to associate words, represented by phoneme strings, with the visual images they described. Recently, \cite{harwath_glass_asru_2015, harwath_nips, Synnaeve14learningwords} introduced models capable of learning the semantic correspondences between raw speech waveforms and natural images at the pixel level. Subsequent works have continued to explore the leveraging of visual information to guide models of speech audio data \cite{harwath_acl_2017, kamper_2017a, chrupala_2017, scharenborg_2018, harwath_icassp_2018, harwath_eccv_2018}, and several papers have begun to investigate the nature of the internal representations learned by these visually-grounded models \cite{drexler_2017, alishahi_2017}. This paper follows the same general theme, but with a different focus. While \cite{drexler_2017} and \cite{alishahi_2017} examined the utility of the intermediate representations of visually-grounded speech models to perform tasks such as speaker, phoneme, and word discrimination, they did not investigate if and how discrete, sub-word units may be emerging within the models. Visually-grounded, self-supervised models such as DAVEnet make relatively few assumptions about how sub-word units should be represented. Therefore, if interpretable sub-word unit structure emerges naturally within the network as a by-product of training, the learned structure could provide a fruitful direction for subsequent research on acoustic unit learning. In this paper, we present experiments that suggest that diphone-like structure is being learned by the intermediate layers of the DAVEnet (Deep Audio Visual Embedding network) audio model \cite{harwath_eccv_2018}. 

\section{The Visually-Grounded Acoustic Model}
\label{sec:davenet}
\begin{figure}[htb]
  \centering
  \centerline{\includegraphics[width=5.5cm]{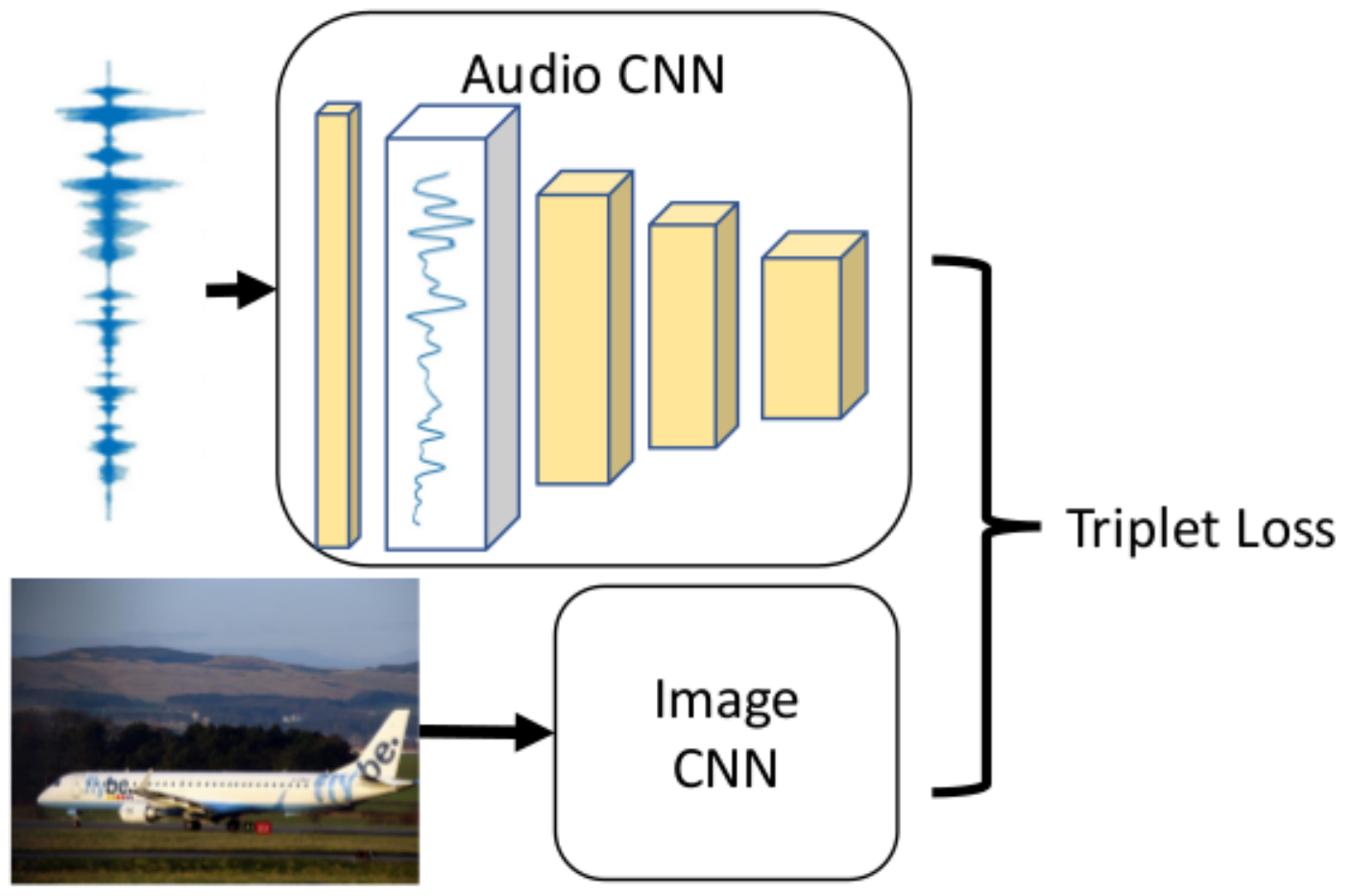}}
\caption{The DAVEnet audio-visual model. We analyze the activation envelope $e[n]$ of the \texttt{conv2} layer.}
\label{fig:davenet_model}
\end{figure}

For our experiments, we leverage the DAVEnet 5-layer speech CNN described in \cite{harwath_eccv_2018}. This model takes as input log Mel-filterbank spectrograms representing a speech signal, and outputs an embedded representation of the speech intended to capture the high-level semantics of the utterance. This is enforced by training the model to associate natural image scenes (encoded with a separate CNN) with spoken captions describing the content of the images (Figure \ref{fig:davenet_model}). Model performance is measured via Recall@10 on an image/caption retrieval task. The speech model is trained from randomly initialized weights, without any traditional linguistic supervision such as word or phonetic transcriptions, a pronunciation lexicon, etc. As in \cite{harwath_eccv_2018}, we train the model on 400,000 image/caption pairs from the Places Audio Caption dataset \cite{harwath_eccv_2018,harwath_nips, places}. We make two small modifications to the model training. First, we employ within-batch semi-hard negative mining \cite{schroff_2015, jansen_icassp_2018}. As in \cite{jansen_icassp_2018}, we blend the semi-hard negative triplet loss with the standard random-sampled triplet loss; for our experiments we simply weight these terms equally. Second, rather than Matchmap-based similarities, we employ global average pooling to the outputs of both networks, and compute their similarity with a dot product. This is mathematically equivalent to the SISA loss described in \cite{harwath_eccv_2018}, but is more computationally efficient for negative mining. Using semi-hard negative mining, we see a boost in Recall@10 from .559 to .641 for image retrieval, and from .506 to .616 for caption retrieval when using a visual model pre-trained on ImageNet \cite{imagenet}.

\section{Experiments}
\label{sec:experiments}
\begin{figure*}[htb]
\centering
\includegraphics[width=.9\textwidth]{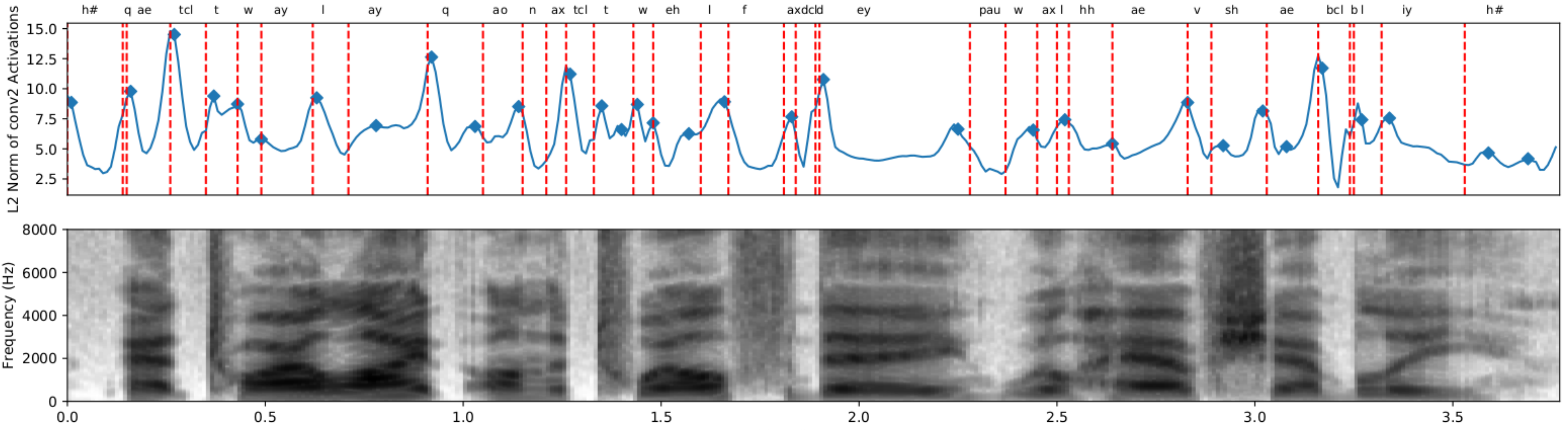}
\caption{The spectrogram for TIMIT utterance \texttt{fisb0\_sx49} (bottom) and its associated $e[n]$ signal (top, blue curve). Peaks found in $e[n]$ shown by blue diamonds, and ground-truth TIMIT phone boundaries denoted by the vertical dashed red lines).}
\label{fig:spectrogram_and_activation_envelope}
\end{figure*}
We focus our analysis on the representations learned by the \texttt{conv2} layer of DAVEnet, which was previously shown by \cite{drexler_2017} to encode more phonetic information than other layers of the network. This makes intuitive sense because its receptive field size (125 ms) more closely corresponds to a typical phonetic segment duration (82 ms on the TIMIT database) than the receptive field size of the \texttt{conv1} layer (25 ms) or the \texttt{conv3} layer (400 ms). When visually examining the outputs of the \texttt{conv2} layer, we observed that the activations tend to oscillate with time (Figure \ref{fig:spectrogram_and_activation_envelope}). This inspired us to investigate the reason for these oscillations in more depth. Given an input spectrogram of $N$ frames, we denote the the activation map output by the second convolutional layer (post-nonlinearity, pre-maxpooling) of DAVEnet as $A \in \mathcal{R}^{Nx256}$, where $A[n,f]$ represents the output of filter $f$ at frame $n$. We compute the activation envelope signal $e[n]$ by taking the L2 norm across all filter channels, i.e. $e[n] = (\sum_{f}A[n,f]^2)^{.5}$. Figure \ref{fig:spectrogram_and_activation_envelope} depicts $e[n]$ and its associated spectrogram for TIMIT \cite{timit} utterance \texttt{fisb0\_sx49}. An interesting property of $e[n]$ is that it is relatively smooth, and exhibits distinct peaks, indicating that there are particular moments in time that trigger strong activity within the layer. These peaks appear to synchronize with phoneme transitions in the spectrogram, which we validate by applying a simple peak-picking algorithm to the envelope signal and measuring the temporal correspondence between these peaks and the ground-truth phonetic boundary annotations for TIMIT. While any standard peak picking algorithm could be used here, we convolve $e[n]$ with a derivative of Gaussian (DoG) filter (whose shape is controlled via a single hyperparameter $\sigma$), i.e. $d[n] = DoG_\sigma[n] * e[n]$. Peaks correspond to positive-to-negative zero crossings in $d[n]$, which are further filtered by a sharpness threshold $\tau$ which compares the maximum slope on the rising edge of a peak to the minimum slope on the falling edge; we keep only those peaks for which the difference between these slopes exceeds $\tau$.

 Our first experiment measures how well the peaks extracted from $e[n]$ correspond to phonetic boundaries on the full test set of the TIMIT corpus, computing precision, recall, and F1 against the ground-truth boundaries. We use the Places audio caption DAVEnet model as-is, and do not do any further training or adaptation on the TIMIT data. We follow \cite{scharenbourg_2010} and use a 20ms tolerance window for boundary detection. We performed a grid search over $\tau$ and $\sigma$, and achieved a maximum F1 of .792 at $\tau=0.15$ and $\sigma=0.5$; but performance was not very sensitive to these exact settings. In Table \ref{tab:bounds}, we compare against several published approaches for blind phone boundary detection. Our method outperforms all of them in terms of F1 score, but does not constitute a fair comparison because our model underwent self-supervised training on the Places audio captions. The key takeaway is the fact that $e[n]$ performs very well as a phone boundary detector despite never being explicitly trained to do so. 
\begin{table}
\small
\centering
\begin{tabular}{c|c|c|c}
\hline
     Algorithm & Precision & Recall & F1 \\
     \hline
     $e[n]$ peaks & .893 & .712 & .792 \\
     \cite{lee_glass_2012} & .764 & .762 & .763 \\
     \cite{michel_2017} & .748 & .819 & .782 \\
     \cite{rasanen_2014} & .740 & .700 & .730 \\
     \hline
\end{tabular}
\caption{Boundary detection on the full TIMIT test. Note that \cite{lee_glass_2012} reflects scores on the training set, not the testing set.}
\label{tab:bounds}
\end{table}

\begin{figure}[htb]
\centering
\centerline{\includegraphics[width=6cm]{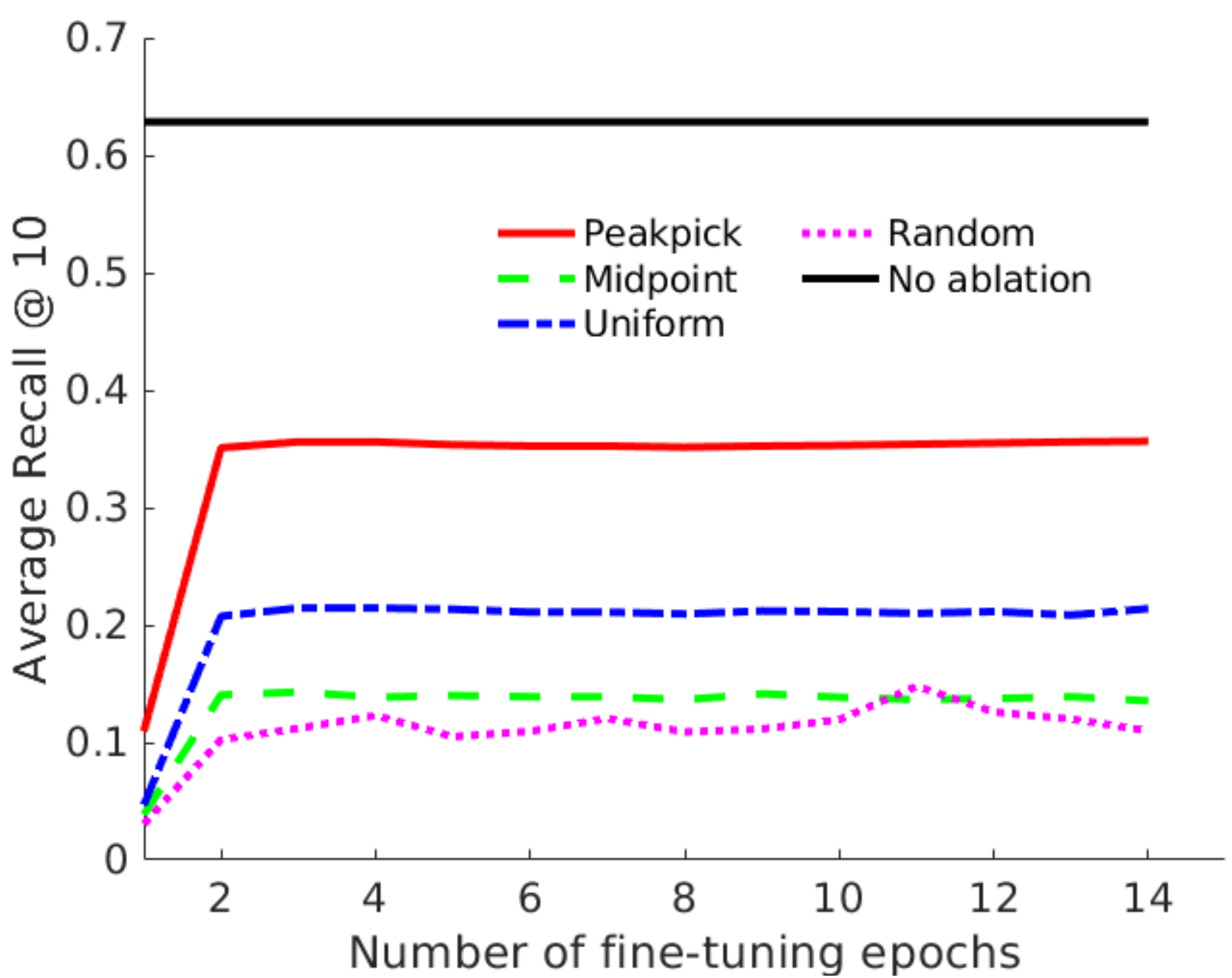}}
\caption{R@10 scores using various ablation methods.}
\label{fig:retrieval}
\end{figure}
Given that some regions of an input spectrogram give rise to peaks in the activation pattern of the \texttt{conv2} layer of DAVEnet, our second experiment examines to what extent the DAVEnet model is leveraging the information contained in these peaks to perform cross-modal retrieval. Here, we fix all of the weights of the DAVEnet model \textit{except} the bias vector of the \texttt{conv3} layer. We then insert an ablation layer between \texttt{conv2} and \texttt{conv3}. This layer computes the peaks in the $e[n]$ signal and then uses them to create a mask matrix $M \in \mathcal{R}^{Nx256}$, where $M[n,f] = 1$ if $e[n]$ has a peak at $n$, and $M[n,f] = 0$ otherwise. The ablated outputs of \texttt{conv2} are then computed as $\hat{A} = M \odot A$, and $\hat{A}$ is fed as input to the subsequent layers of the network. Because the ablation layer changes the magnitude of the summed input seen by each neuron in the \texttt{conv3} layer, we fine-tune only the bias of this layer on the ablated outputs from \texttt{conv2}, using the image and caption ranking objective described in Section \ref{sec:davenet}. By comparing the retrieval recall scores achieved by the ablated network against those of the original model, we can infer to what extent the convolutional filters of the already-trained DAVEnet model have learned to focus on the \texttt{conv2} activation peaks, as opposed to other regions of the \texttt{conv2} output. Figure \ref{fig:retrieval} displays the average of the image-to-caption and caption-to-image recall @ 10 scores as a function of the number of fine-tuning epochs. The horizontal black line represents the recall score when no ablation is used (0.629), and the red solid line shows the score achieved when using the peak-picking based ablation. After a single epoch of fine-tuning, the R@10 score rebounds from 0.117 to 0.351, where it remains constant. We compare the peak-picking ablation against uniform sampling, random sampling, and sampling the midpoint frame between each consecutive pair of $e[n]$ peaks. For uniform and random sampling, we keep the number of ablated frames constant for each utterance; if a given utterance of length $N$ was found to have $N_p$ peaks in its $e[n]$ signal, then we retain $N_p$ uniformly-spaced or randomly sampled \texttt{conv2} activation frames across the utterance. On average, 1 out of every 11.84 frames was found to be a peak, meaning that 91.6\% of the \texttt{conv2} output frames were set to zero. In Figure \ref{fig:retrieval}, we see that while all ablation methods suffer a loss in retrieval accuracy, the peak-picking ablation model still achieves 60\% of the performance of the non-ablated model. All other methods fare worse, indicating that the filters of the DAVEnet audio model have learned to leverage the $e[n]$ peaks for word discrimination much more than other parts of the speech signal.

\begin{figure}[htb]
  \centering
  \centerline{\includegraphics[width=6cm]{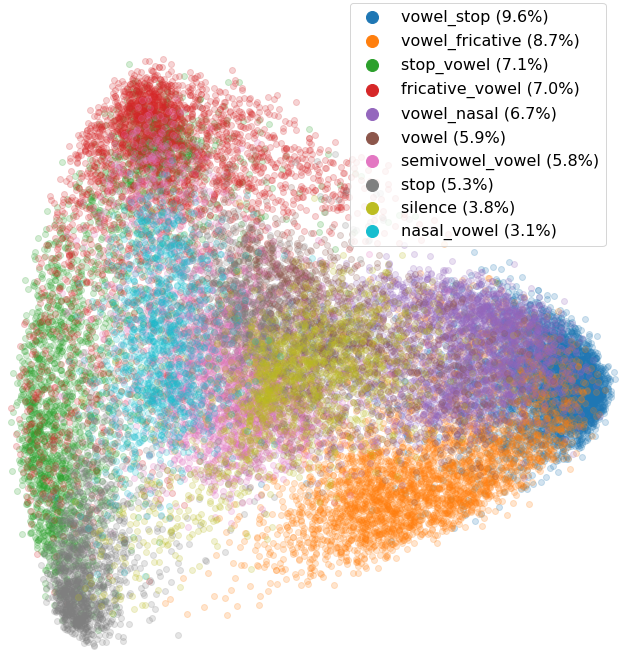}}
\caption{PCA analysis of $e[n]$ peaks extracted from the TIMIT full test set.} 
\label{fig:pca}
\end{figure}

Thus far, we have shown that the \texttt{conv2} layer of the DAVEnet audio model is highly sensitive to specific regions of an input spectrogram, that these regions are especially informative for inferring the semantics (and thus the lexical content) of an utterance, and that these regions tend to occur at the transition point between two phones. Our last experiments investigate the geometry of the embedding space in which these activation peaks reside. We first extracted a total of 39,871 peaks for the 1,344 utterances comprising the TIMIT complete test set. We represent each peak with its corresponding 256-dimensional embedding vector produced by the \texttt{conv2} layer of DAVEnet. We then assign a label to each peak according to the ground-truth sequence of phones that fall within a 40 ms window around the peak. We follow a similar scheme to the 39-phone mapping for TIMIT, but map stop closures to their associated stop phoneme instead of silence. Under this mapping, we found that approximately 18.1\% of the peaks fell within a single phone segment, 76.5\% of peaks captured a diphone boundary, and 5.3\% of the peaks overlapped three phones. In addition to a phonetic label for each peak, we also derive a manner label by mapping each phone to its associated broad manner class (vowel, stop, nasal, fricative, semivowel, affricate, flap, and silence). We projected the peak embeddings down to 2 dimensions using PCA, and plot the peaks corresponding to the 10 most frequently occurring broad manner class labels in Figure \ref{fig:pca}. We can see that peaks belonging to the same underlying manner class cluster together quite well. Furthermore, we observe that CV, V, and VC syllable structure is captured by the first principal component; vowel\_stop, vowel\_fricative, and vowel\_nasal peaks are concentrated on right-hand side of the space, while stop\_vowel, fricative\_vowel, and nasal\_vowel reside on the left-hand side of the space.

\begin{figure}[htb]
\begin{minipage}[b]{.48\linewidth}
  \centering
  \centerline{\includegraphics[width=3.5cm]{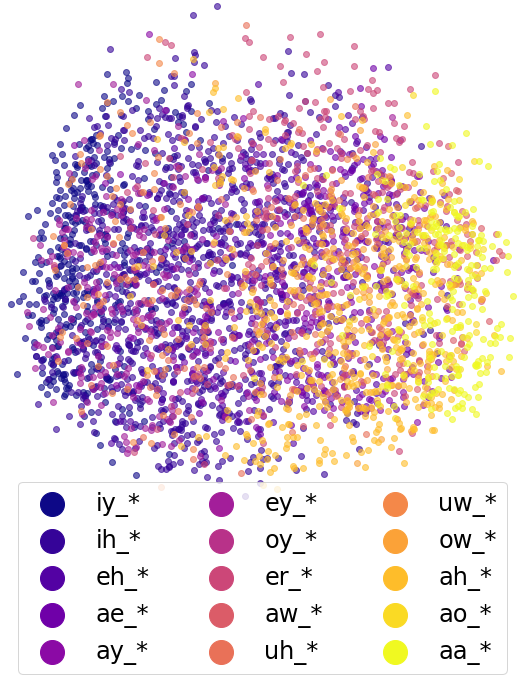}}
\end{minipage}
\hfill
\begin{minipage}[b]{0.48\linewidth}
  \centering
  \centerline{\includegraphics[width=3.5cm]{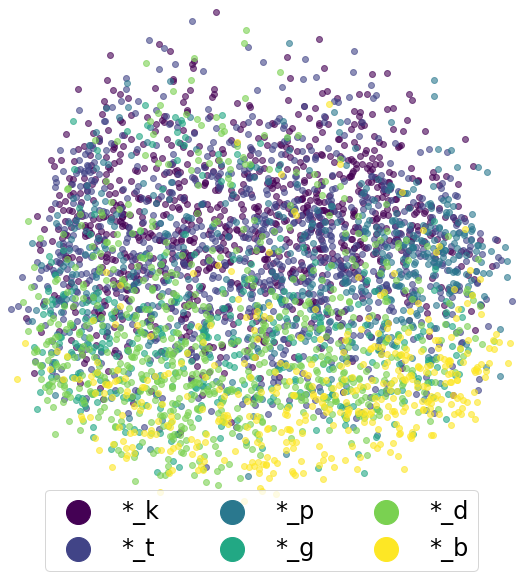}}
  \vspace{.7cm}
\end{minipage}
\caption{PCA analysis of the peaks corresponding to vowel-stop transitions.}
\label{fig:vowel_stop_pca}
\end{figure}
\begin{figure}[htb]
  \centering
  \centerline{\includegraphics[width=6cm]{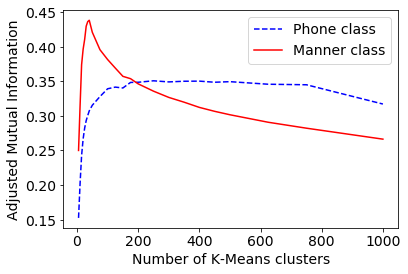}}
\caption{Adjusted mutual information between K-means clustering of peaks and their underlying phone and manner class sequences.} 
\label{fig:ami}
\end{figure}
In Figure \ref{fig:vowel_stop_pca}, we examine the properties of peaks belonging to the vowel\_stop class in more detail. We compute a second PCA transform specific for this class, and plot the associated peaks along their first two principal components. In the left-hand scatter plot, the peaks are color coded according to the vowel in the left context of the peak; in the right-hand plot, they are color coded according to their right-hand stop context. Broadly speaking, the first principal component seems to select for frontness of the vowel, while the second component captures the voicing of the stop consonant. 

Our qualitative analysis suggests that the discovered peaks cluster by manner class at a coarse scale, and by phonetic identity at a finer scale. We quantify this by clustering the peak vectors using K-means, and computing the adjusted mutual information (AMI) \cite{vinh_epps_bailey_2010} between the clustering output and the underlying phone and manner class sequences for each peak (Figure \ref{fig:ami}). AMI is maximized for the manner class label sequences at $K=40$ clusters and steadily falls off as more clusters are specified, while the AMI for phonetic labels plateaus between 200 and 700 clusters.

\section{Concluding Discussion}
In this paper, we investigated the encoding of sub-word information in the \texttt{conv2} layer of the DAVEnet visually-grounded speech model. We observed that the magnitude of the activations within this layer tend to exhibit local maxima at diphone transitions, which we quantified by using these maxima to detect phone boundaries on TIMIT. Furthermore, we performed ablation experiments for an image/caption retrieval that suggested that the DAVEnet audio model leverages the $e[n]$/``diphone'' peaks more than other regions of the signal for the purpose of word recognition. Finally, we examined the geometry of the space occupied by the peak embedding vectors and found the emergence of clusters of diphone units which share broad phonetic manner class membership; within these clusters, different dimensions appear to correlate with distinctive features such as vowel frontness or stop voicing.  In our future work, we plan to further explore the topic of leveraging visually-grounded acoustic models to discover discrete, pseudo-linguistic units. We would like to explicitly incorporate mechanisms into DAVEnet for inferring a discrete, compositional hierarchy of interpretable phone-like, syllable-like, word-like, phrase-like, etc. units that could provide a rich account of a spoken language in a self-supervised fashion. Finally, we believe that our ablation analysis in Section \ref{sec:experiments} points the way towards a non-linear downsampling scheme that would enable acoustic observation sequences to be more closely aligned with phone or character sequences, which could find application in supervised ASR systems.

\vfill
\pagebreak
\footnotesize
\bibliographystyle{IEEEbib}
\bibliography{strings,refs}

\end{document}